\documentclass{article}
\PassOptionsToPackage{numbers, compress}{natbib}

\usepackage[final]{nips_2017}
\usepackage{booktabs}      
\usepackage{amsfonts,amsmath,amssymb}
\usepackage{nicefrac} 
\usepackage{microtype}
\usepackage{graphicx}	
\usepackage{todonotes}

\begin{document}

\title{Learning to Plan Chemical Syntheses}

\author{Marwin Segler$^{\clubsuit,\nabla}$  \qquad  Mike Preuss$^H$ \\
$^\clubsuit$Institute of Organic Chemistry\\
$^\nabla$Center for Multiscale Theory and Computation\\
$^H$Department of Information Systems\\
Westf\"alische Wilhelms-Universit\"at M\"unster\\
\texttt{\{marwin.segler,mike.preuss\}@wwu.de} \\
\And
Mark P. Waller$^{\spadesuit\hbar}$\\
$^\spadesuit$Department of Physics\\
$^\hbar$International Center for Quantum\\ and Molecular Structures\\
Shanghai University\\
\texttt{waller@shu.edu.cn} \\
}

\maketitle
\begin{abstract}
From medicines to materials, small organic molecules are indispensable for human well-being. To plan their syntheses, chemists employ a problem solving technique called retrosynthesis. In retrosynthesis, target molecules are recursively transformed into increasingly simpler precursor compounds until a set of readily available starting materials is obtained. 
Computer-aided retrosynthesis would be a highly valuable tool, however, past approaches were slow and provided results of unsatisfactory quality. 
Here, we employ Monte Carlo Tree Search (MCTS) to efficiently discover retrosynthetic routes. MCTS was combined with an expansion policy network that guides the search, and an ``in-scope'' filter network to pre-select the most promising retrosynthetic steps. These deep neural networks were trained on 12 million reactions, which represents essentially all reactions ever published in organic chemistry. Our system solves almost twice as many molecules and is 30 times faster in comparison to the traditional search method based on extracted rules and hand-coded heuristics. Finally after a 60 year history of computer-aided synthesis planning, chemists can no longer distinguish between routes generated by a computer system and real routes taken from the scientific literature. We anticipate that our method will accelerate drug and materials discovery by assisting chemists to plan better syntheses faster, and by enabling fully automated robot synthesis.

\end{abstract}

Retrosynthetic analysis is the canonical technique to plan the synthesis of organic small molecules, for example drugs, agro- and fine chemicals, and is part of every chemist's curriculum.\cite{clayden2008organic,bruckner2014} In retrosynthesis, a search tree is built by ``working-backwards'', analysing molecules recursively and transforming them into simpler precursors until one obtains a set of known or commercially available building block molecules.\cite{corey1989logic} The transformations are reversed chemical reactions: A chemist can take the plan, and execute it in the lab in the forward direction to synthesize the target compound. Retrosynthetic analysis is a formidable intellectual exercise that demands broad and deep chemical knowledge. Its central element requires a combination of creativity and a pattern recognition process, in which functional groups (patterns of atoms and bonds) of the target molecule are matched to transformations.\cite{todd2005computer}

Transformations are derived from successfully conducted series of similar reactions with analogous starting materials, and often named after their discoverers.\cite{kurti2005strategic} At each retrosynthetic step, out of thousands of known transformations in modern chemistry, chemists intuitively rank the most promising transformations highly, and do not even actively think about the unreasonable ones.\footnote{In cognitive science, this can is explained via ``System 1'' and ``System 2'' in dual process theory.\cite{evans2012dual}} Unfortunately, when a transform is applied to a new molecule, there is no guarantee that it will generalize and the corresponding reaction will actually proceed in the expected way.\cite{collins2013robustness} A molecule failing to react in the way predicted by the transform is called 'out of scope'. This can be  due to steric or electronic effects, an incomplete understanding of the reaction mechanism, or conflicting reactivity in the molecular context, which may hamper a reaction and lead to chemo-, regio- or stereoselectivity issues. Predicting which molecules are 'within scope' can be challenging even for the best human chemists, who perform synthesis planning using knowledge and mechanistic reasoning acquired during long years of study and the many intuitive models the chemical community developed over the last 250 years, in combination with extensive literature research for every transform step, which may take hours.\cite{collins2013robustness, corey1989logic}

Computer-assisted synthesis planning (CASP) could help chemists to find better routes faster, and is a missing component in virtual de novo design and robot systems performing \textit{molecular design-synthesis-test} cycles.\cite{ley2015organic,hartenfeller2011enabling,schneider2016novo,segler2017generating} 
To perform CASP, the knowledge that humans ``simply learn'' has to be transferred into an executable program.\cite{vleduts1963concerning,todd2005computer,szymkuc2016computer,cook2012computer,ihlenfeldt1996computer} 
Despite 60 years of research and industrial-scale backing, attempts to formalize chemistry by manual encoding by experts have not convinced synthetic chemists.\cite{ihlenfeldt1996computer,ugi1994models,kayala2011learning,kayala2012reactionpredictor,minsky1974framework,fickdiss} Approaches to algorithmically extract transformations from reaction datasets\cite{bogevig2015route,law2009route,christ2012mining} on a purely symbolic level were criticized for their high amount of noise and lack of ``chemical intelligence''.\cite{cook2012computer,szymkuc2016computer} 
However, we recently showed that deep neural networks trained on large reaction datasets are capable of ranking extracted symbolic transformations, and can learn to avoid reactivity conflicts.\cite{neural-symbolic} This allows to mimic the expert's intuitive decision-making.\cite{neural-symbolic}  

\begin{figure}[htb]
\begin{center}
\includegraphics[width=1.0\linewidth]{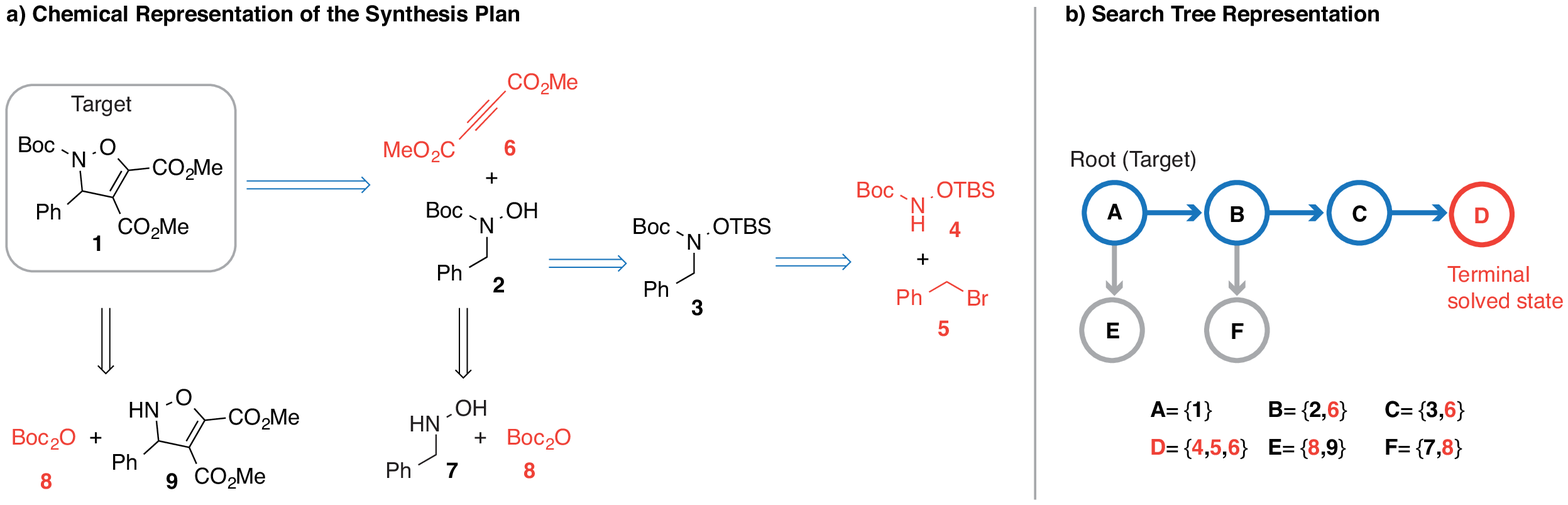}
\caption{Translation of a) the retrosynthetic route representation (conditions omitted)\cite{olga} to b) the search tree representation. The target molecule can be solved if it can be deconstructed to a set of readily available building blocks (marked red).}
\label{fig:retro}
\end{center}
\end{figure}

In retrosynthetic analysis, the sequential transformation of the target molecule gives rise to a structure called search tree. 
Figure \ref{fig:retro} translates the traditional chemist's representation to a search tree representation.\cite{szymkuc2016computer} The nodes in the tree represent the \textit{synthetic position}. Positions are sets of molecules; they contain all precursors needed to make the molecules of the preceding positions all the way down to the tree's root node (which contains the target molecule). Branches in the search tree then correspond to complete routes, which allows to compare and rank different routes.

Search in retrosynthesis poses a significant challenge: The branching factor (number of applicable transformations per step) can be quite large and range from 80 for modest systems\cite{szymkuc2016computer} up to more than 40000 (see below). Even though the search depth rarely exceeds more than 15-20 steps, the search tree explodes combinatorially if solutions are explored exhaustively. 
In the past, heuristic best first search (BFS) has been employed for synthesis planning.\cite{szymkuc2016computer} Heuristic functions determine position values, which can guide the search into promising directions. Unfortunately, unlike in chess, it is difficult to define strong heuristics in chemistry for two reasons: First, chemists tend to disagree about good positions, that is how easy it is to make a molecule.\cite{boda2007structure,ertl2009estimation} Second, the value of a position depends highly on the availability of suitable precursors. \cite{ihlenfeldt1996computer,szymkuc2016computer} Even complex molecules can be easily made in a few steps if there are readily available building blocks. Therefore, one cannot reliably estimate the value of a synthetic position without completely ``playing the molecules until the end of the game''.

Recently, Monte Carlo Tree Search (MCTS) has emerged as a highly general search technique, which is particularly suited for sequential decision problems with large branching factors without strong heuristics, such as games or automated theorem proving.\cite{kocsis2006bandit,browne2012survey,farber2016monte} 
These properties suggest that MCTS could be effective in synthesis planning as well. 
MCTS builds a search tree by iterating four distinct phases, see Figure 2a. During \textit{selection} (1), the most important position to analyse next is selected according to its expected value and the confidence in that value. \textit{Expansion} (2) performs one step of retroanalysis on the chosen position by adding possible precursor positions. One of these precursor positions is then selected for evaluation by a \textit{rollout} (3). A rollout is a Monte Carlo simulation, in which random search steps are performed without branching until a solution has been found by transforming the position into building blocks or a maximum depth is reached. 
During \textit{update} (4), the position values in the tree are updated to reflect the outcome of the rollout. As the tree grows and many iterations are run, the position values become more accurate, and eventually converge to the optimal solution. 

For the game of Go, Silver et al. elegantly demonstrated with the AlphaGo system that MCTS can be augmented with neural network \textit{policies} $p(t|s)$, which predict the probability of taking move (applying transform) $t$ in position $s$, and are trained to predict the winning move based on grandmaster games. Such policies were used to inform selection and rollout instead of taking random moves.\cite{silver2016mastering,maddison2014move,storkey2015training}

In this work, inspired by the recent progress in Go culminating in AlphaGo,\cite{silver2016mastering} we combine three different neural networks (3N) together with MCTS to perform chemical synthesis planning (3N-MCTS). The first neural network (expansion policy), trained on how molecules were made in the past, transfers its knowledge to novel molecules. It guides the search into promising directions by proposing a restricted number of possible, automatically extracted transformations. A second neural network then predicts if the proposed reactions are actually feasible/in scope. Finally, to estimate the position value, transformations are sampled from a third neural network (policy) during rollout. The neural networks were trained on essentially all reactions published in the history of organic chemistry.

\begin{figure}[ht!]
\begin{center}
\includegraphics[width=\textwidth]{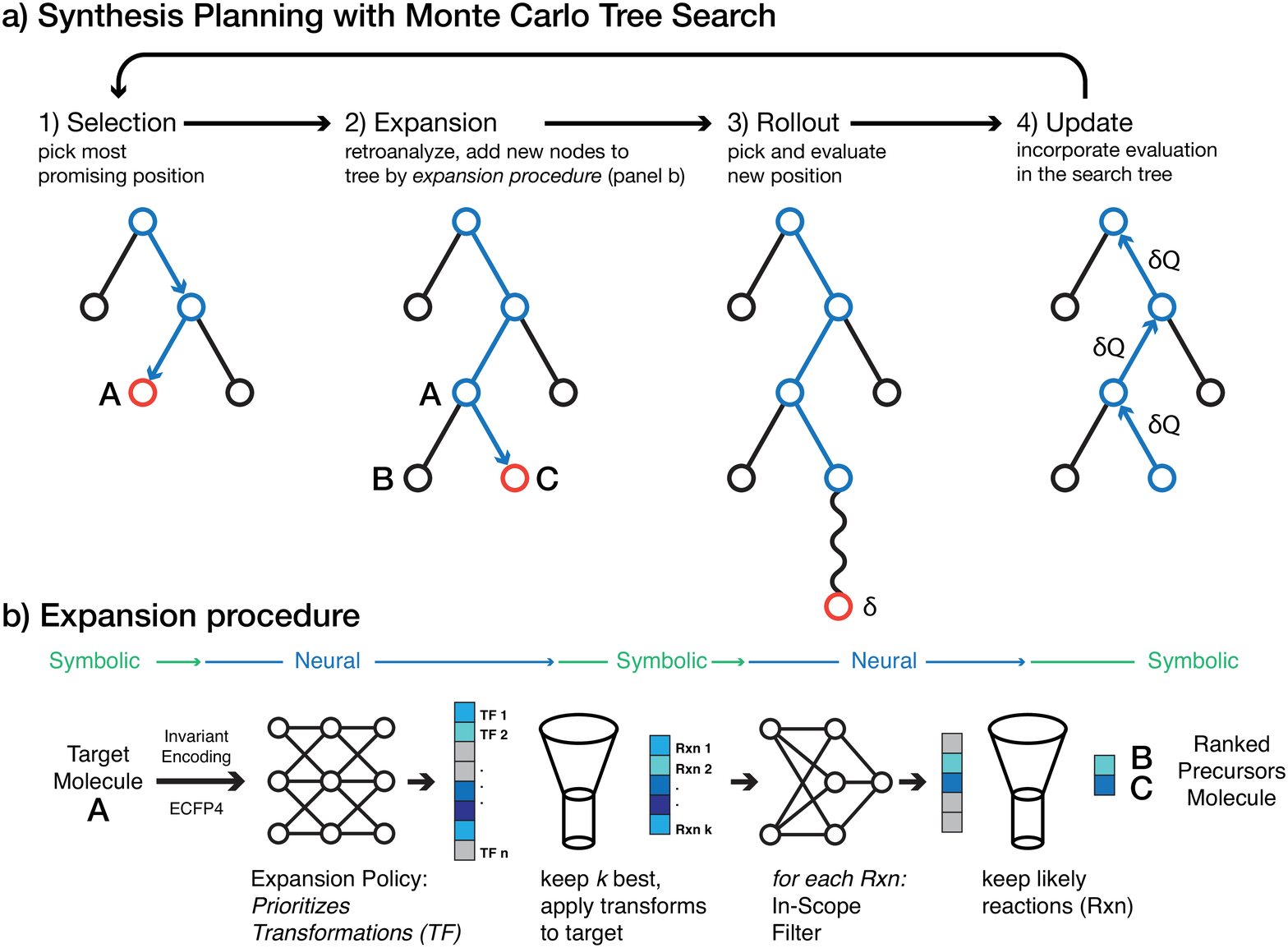}
\caption{a) Monte Carlo Tree Search(MCTS) performs search by iterating over 4 phases: In the selection phase 1), the most urgent node for analysis is chosen based on the current position values. In phase 2) this node may be expanded by processing the molecules of the position A with the Expansion procedure (b), which leads to new positions B,C which are added to the tree. Then, the most promising new position is chosen, and a rollout phase 3) is performed by randomly sampling transforms from the rollout policy until all molecules are solved or a certain depth is exceeded.  In the update phase 4), the position values are updated in the current branch to reflect the result of the rollout. 
b) Expansion procedure: In the step, the molecule $m_t$ to retroanalyse is fed into the policy network to predict the best transforms. Then, only the top $k$ transforms are applied to the molecule, which yields complete reactions (Rxn). For each of these reactions, reaction prediction is performed using the transition classifier. Improbable reactions are then filtered out, which leads to the list of admissible actions and corresponding precursor positions B,C. (TF = Transformation, Rxn=Reaction, ECFP = Extended Connectivity Fingerprint)}
\label{fig:mcts}
\end{center}
\end{figure}

\subsection*{Automatic extraction of transformation rules from reaction datasets}

We extracted transformation rules from 12.4 million single step reactions from the Reaxys\cite{reaxys} database using our previously reported protocol, which took approx. 3 hours on a laptop.\cite{neural-symbolic} The database contains the bulk of organic and organometallic reactions published since the inception of the field in 1771 until today.\cite{reaxys} 
Two sets of rules were extracted: The rollout set comprises rules which contain the atoms and bonds changed in the course of the reaction (the reaction center), and the first degree neighbouring atoms. Here, all rules which occurred at least 50 times in reactions published before 2015 were kept. For the expansion rules, a more general rule definition was employed. Here, only the reaction center is extracted. Rules occurring at least 3 times were kept. 
The two sets encompass 17,134 and 301,671 rules, and cover 52\% and 79\% of all post-2014 chemical reactions, respectively.
\footnote{
It has to be noted that even when taking all pre-2015 rules without count restriction into account, only 82\% of the reactions published after 2014 are covered. The missing 18\% are novel reaction types. This highlights the success of chemists inventing novel methodologies, but also implies that eventually a retrosynthesis system should be also able to discover novel reactions on its own.
}

\subsection*{Training the Expansion Policy and Rollout Policy Network}
Rule extraction associates each reaction, and thus each product with a transformation rule. 
As previously reported, this allows to train neural networks as policies to predict the best transformations given the product, or in other words, the best reactions to make the product.\cite{neural-symbolic} Importantly, such neural networks also learn about the context in which the reactions can occur (functional group tolerance).\cite{neural-symbolic} 

As the expansion policy, we employed a deep highway network with exponential linear unit (ELU) nonlinearities.\cite{srivastava2015training,clevert2015fast}  
To assess its generalisation ability, we performed a time-split strategy. For training, only reactions published before 2015 were used, while for validation and testing, data from 2015 and later was selected. Time-split validation has been shown to give a more realistic assessment of predictive performance than other splitting strategies, such as random splitting.\cite{sheridan2013time} 
Table 1 shows the metrics for the expansion policy. The neural network predicts the correct solution out of 301,671 transformations with an accuracy of 31\%, which is reasonable. It has to be noted that there are almost always many feasible ways to make a molecule. Top10 and top50 accuracies of 63.3\% and 72.5\% indicate that the correct transformations are generally ranked highly.

Beyond the top50 predicted results, the accuracy only increases marginally. This observation allows to drastically reduce the branching factor. During search tree expansion, we restrict the possible transformations to a maximum of 50. Additionally, we sum up the probabilities of the predicted actions starting from the highest ranked transformation. When the cumulative probability reaches 0.995, we stop further expansion, even if less than 50 actions have been expanded. This allows the system to focus on highly likely transformations in cases where only few good options exists to make a molecule, for example acyl chlorides or Grignard reagents.\footnote{Additionally, this entails that the NP-complete subgraph isomorphism problem, which determines if the corresponding rule can be applied to a molecule and yields the next molecule(s), needs only to be solved for a subset of the rules.} We observed that the reactions in this reduced top50 are almost exclusively reasonable, and often variations of the correct prediction. For example, a Heck reaction can usually be conducted with bromide, iodide or triflate as the leaving group.

The rollout policy network, which is a neural network with one hidden layer, is trained in the same way as the expansion policy. It uses a set of 17,134 rules, which leads to a lower coverage, yet it needs just 10~ms to make a prediction, in contrast to 90~ms for the expansion policy (see Table 1). 
The rationale for using two different rulesets is to have a powerful, but slow policy to select the best candidate transformations for expansion, and a fast roll policy to estimate the position values.\cite{silver2016mastering}

\begin{table}[htb]
\caption{Metrics for the three Supervised Neural Network Policies}
\begin{center}
\begin{tabular}{lrrrrrr}
\toprule
Policy&  \# rules& Coverage& Matching rules/mol.\footnote{This corresponds to the branching factor. Calculated on 500 randomly selected molecules first synthesized after 2015.} &  Acc&   top10Acc&   top50Acc\\
\midrule
Expansion& 301,671 & 0.79	& 46,175 &	0.310&  0.633&	0.725\\
Rollout	&  17,134 & 0.52 &   321 &   0.501&	0.891&	0.964\\
\bottomrule
\end{tabular}
\end{center}
\label{tab:metrics}
\end{table}%

\subsection*{In-Scope Filter Network}
After the search space has been narrowed down by the expansion policy to the most promising transformations, we need to predict whether the corresponding reactions will actually work for a particular molecule. We trained a deep neural network as a binary classifier to predict whether the reactions corresponding to the transformations selected by the policy network are actually feasible.\cite{marcou2015expert,Polishchuk2017} The classifier has to be trained on successful and failed reactions. Unfortunately, failed reactions are only seldom reported and not contained in reaction databases. Therefore, starting from reported reactions, negative reactions, for example with incorrect regio- and chemoselectivity were generated using an established protocol\cite{carrera2009machine,segler2017modelling,coley2017prediction} together with negative sampling. Using this data augmentation strategy, we generated 100 million negative reactions from reactions prior to 2015 for training and 10 million after 2014 for testing. As positive cases, reported reactions from these periods are used. The classifier only receives the product and the reaction fingerprint as its input -- reaction conditions or additional reagents or catalysts are not considered. The classifier thus only serves as rapid filter to predict if a reaction will generally work or not. 
On the test set, the classifier achieves an area under the Receiver Operation Characteristic (ROC) curve of $0.99$, and the area under the precision-recall curve of $0.94$, which indicates good performance.\cite{murphy2012machine}

\subsection*{Expansion procedure}

The expansion policy network and the in-scope filter network are then combined to a pipeline (Figure \ref{fig:mcts} b)).  When a position $A$ is to be analyzed, each molecule of the position is fed into the policy network. Then, the transformations with the highest scores are applied to the molecule, which yields the possible precursors and thus full reactions. These reactions are submitted to the in-scope filter, where only transformations and precursors corresponding to positively classified reactions are kept. They represent the ``legal moves'' available in position $A$.

\subsection*{Integrating Neural Networks and MCTS for Synthesis Planning}
The expansion procedure and the rollout policy are then incorporated in the respective phases of a Monte Carlo Tree Search algorithm to form our 3N-MCTS ansatz. The four phases are iterated to build the search tree.

\paragraph{1. Selection}
In the first 3N-MCTS phase, starting at the root node (the target molecule) of the search tree, the algorithm sequentially selects the most promising next position within the tree until a leaf node is reached (Fig. \ref{fig:mcts}a). The algorithm balances the selection of high-value positions and unexplored positions.  If a leaf nodes is visited for the first time, it is directly evaluated by a rollout. If it is visited for the second time, it is expanded by processing via the expansion policy.

\paragraph{2. Expansion}
In the expansion, the possible transforms determining the follow-up positions of the current position are selected by applying the expansion procedure. The predicted follow-up positions are added to the tree as children of the leaf node, and the most promising position is selected for rollout. 

\paragraph*{3. Rollout}

In the rollout phase, first, the status of the position is checked. If the position is already solved, the algorithm directly receives a reward greater than 1.\cite{winands2008monte} Non-terminal states are subjected to a rollout, where actions are sampled from the rollout network recursively, until the state has been deconstructed into building blocks or a maximal depth is reached.

\paragraph*{4. Update}
If a solution has been found during rollout, a reward of 1 is received. Partial rewards are given if some but not all molecules in the state are solved. If no solution was found, a reward of --1 is received. 
In the last phase, the tree is updated to incorporate the achieved reward by updating the position values. 

These four phases of 3N-MCTS are iterated until a time budget or maximal iteration count is exceeded. Finally, to obtain the synthesis plan, we repeatedly select the retrosynthetic step with the highest value until a solved position is reached, or a maximum depth has been exceeded, in which case the problem is unsolved.

\section*{Evaluating the performance characteristics of 3N-MCTS}

To evaluate the performance of 3N-MCTS, we compare our algorithm to the state of the art search method, which is Best-First Search (BFS) with the hand-coded SMILES$^\frac{3}{2}$ heuristic cost function, as described by Grzybowski and coworkers (heuristic BFS).\cite{szymkuc2016computer} 
This function assigns the lowest cost to reactions that split up the molecule in equally sized parts. Additionally, we perform BFS with the cost calculated by the policy network (neural BFS). 
All algorithms use the same set of automatically extracted transformations. The evaluation is performed in a time-split way: The models were trained only on data published before 2015. As test data, only molecules first reported after 2014 are considered, which were not contained in the training data.

The different algorithms were provided with a set of target molecules, the algorithms had to find a synthesis route to known building blocks. The building blocks have to be selected before the search is started, and could be molecules on stock in the lab, known in the literature, or commercially available chemicals. Herein, we use a set of 423,731  molecules, containing 84,253 building blocks from three major chemical suppliers (SigmaAldrich, AlfaAesar and Acros),\footnote{obtained from the ZINC database http://zinc15.docking.org/catalogs/subsets/building-blocks/} and 339,478 molecules from the Reaxys database, which have been used as reactants at least 5 times before 2015. A researcher may choose different sets of building blocks for each search, e.g. first try to find the solution to a problem with molecules on stock in the lab, and afterwards consider additional molecules from chemical suppliers.

\begin{figure}[htb]
\begin{center}
\includegraphics[width=0.9\textwidth]{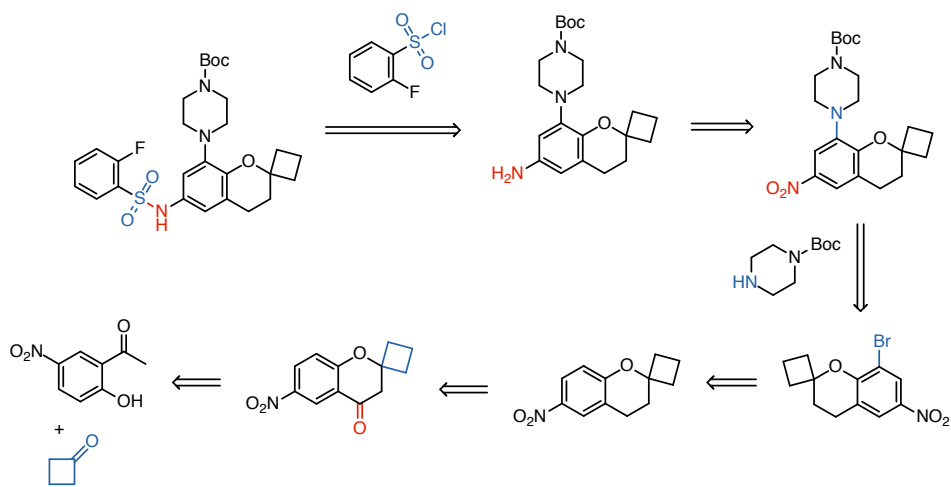}
\caption{An exemplary 6 step synthesis route for an intermediate in a drug candidate synthesis published first in 2015.\cite{nirogi2015design} It is identical to the published route\cite{nirogi2015design} and was found by our algorithm autonomously within 5.4 s. The affected functional groups in each step are marked blue or red.}
\label{fig:example}
\end{center}
\end{figure}

Figure \ref{fig:example} shows an exemplary 6-step route for an intermediate of a drug candidate synthesis reported in 2015, which has been found by our algorithm in 5.4 s. It matches the published route.\cite{nirogi2015design}

\subsection*{Quantitative evaluation}

\begin{figure}[htb]
\begin{center}
\includegraphics[width=0.8\textwidth]{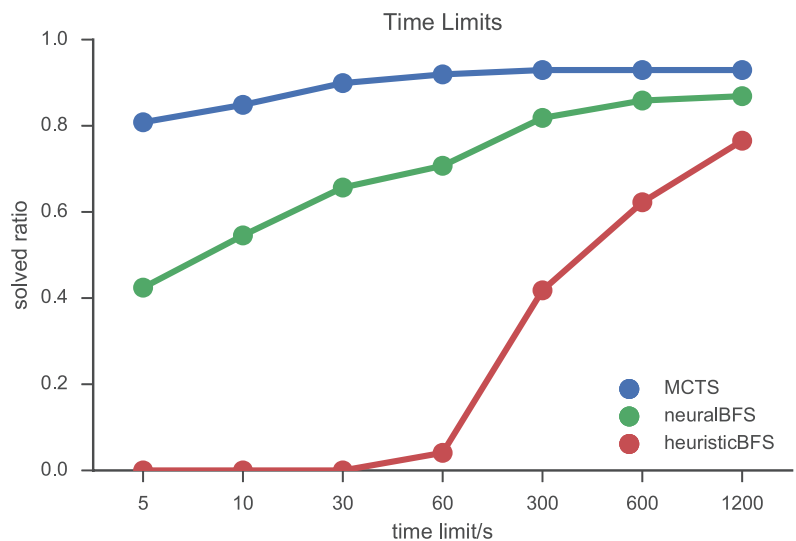}
\caption{Influence of the time per query (max. steps = 100000) on performance.}
\label{fig:performance}
\end{center}
\end{figure}

Surprisingly, in the past, neither hand-coded nor automatically extracted retrosynthetic systems have ever been validated in a statistical way. 
To obtain a set of target molecules that contain different scaffolds, first all molecules reported after 2014 were clustered using the extended connectivity fingerprint (ECFP6)-based Butina algorithm\cite{butina1999unsupervised}. 
Then, 497 target molecules were randomly selected from amongst the different cluster cores. This test set was employed to quantitatively assess the performance characteristics of the different search algorithms by finding synthesis routes to known building blocks (see Figure \ref{fig:performance})
 
MCTS already solves more than 80\% of the test set with a time limit of 5 seconds per target molecule, compared to 40\% with neural BFS and 0\% for heuristic BFS. MCTS solved 92\% of the test set with a limit of 60~s per molecule, whereas neural BFS solved 71\%, and heuristic BFS solved 4\%. Even at much longer runtimes of 20 minutes per molecule, heuristic and neural BFS are not able to compete with MCTS. Provided with infinite runtime, however, the algorithms will converge to the same performance.

To determine which components of MCTS are responsible for its superior performance, we compared MCTS against several related search algorithms (see Table \ref{tab:exp}) at a runtime limit of 3$\times$300~s (three restarts).
MCTS in conjunction with the expansion policy network solved the highest number of retrosynthetic targets. On average, MCTS required the least amount of time per molecule to find a solution (entry 1). 
Plain Monte Carlo search (MC) randomly selects transforms using the expansion policy network, without building a tree. MC (entry 2) solved 89.54\% 
UCT is an MCTS variant which uses the expansion policy network only to narrow down the possible transforms, but not to guide the search via the predicted probability of the transform. This way, 87.12\% of the test set is solved. BFS using a cost function based on the expansion policy network only solved 84.24\%, highlighting the importance of rollouts. The traditional approach, BFS with a hand-designed heuristic cost function, solves only 45.6\% of the test set, and needs 433.4 s on average to find a solution.
These results suggest that all components of 3N-MCTS (building a tree, reducing the branching factor via the expansion policy, guiding the search with the expansion policy, and using rollouts) contribute to its superior performance.

\begin{table}[htb]
\caption{Experimental Results}
\begin{center}
\begin{tabular}{lllrr}
\toprule
Entry & Search Method \qquad & Policy$^a$ \qquad\qquad  & \% solved  & \qquad  time (s/mol)  \\
\midrule
1 & MCTS & Neural       & 95.24 $\pm$0.09 &  13.0\\
2 & MC   & Neural       & 89.54 $\pm$0.59 & 275.7\\
3 & UCT  & Neural       & 87.12 $\pm$0.29 &  30.4\\
4 & BFS  & Neural       & 84.24 $\pm$0.09 &  39.1\\
5 & BFS  & SMILES$^\frac{3}{2}$     &    55.53 $\pm$2.02     &  422.1       \\
\bottomrule
\end{tabular}
\end{center}
\scriptsize{Time budget $300~s$ and $100,000$ iterations for MCTS or $300~s$ and $100,000$ expansions for BFS, per molecule. $3$ restarts were carried out. $^a$In BFS, this is the cost function.}
\label{tab:exp}
\end{table}%

\subsection*{Double-blind AB tests}

The central criticism of retrosynthesis systems has been that the proposed routes often contain what chemists immediately recognize as chemically unreasonable steps. Therefore, to assess the quality of the solutions we conducted two AB tests, in which 45 graduate level organic chemists from two world-leading organic chemistry institutes in China and Germany had to choose one of two routes leading to the same molecule based on personal preference and synthetic plausibility. The tests were double blind, meaning that neither participants nor conductors were aware of the origin of the routes. The test molecules were selected randomly from a set of drug-like compounds first published after 2014.

\begin{figure}[htb]
\begin{center}
\includegraphics[width=1.0\textwidth]{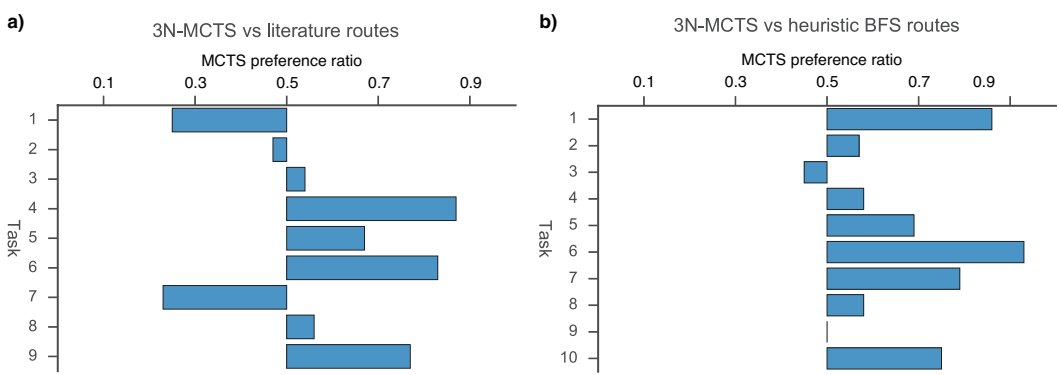}
\caption{a) Chemists did not prefer literature routes over routes found by MCTS b) Chemists significantly prefer routes by 3N-MCTS  over routes by heuristic BFS without policy network and in-scope filter. As ratio above 0.5 indicates that more than 50\% of participants preferred the MCTS solution.}
\label{fig:abtest}
\end{center}
\end{figure}

In the first test, the participants had the choice between a route reported by expert chemists in the literature, and a route generated by our 3N-MCTS algorithm for the same target molecule. Routes to nine different target molecules were offered. Routes towards the same molecule were required to have the same number of steps.

\begin{figure}[h]
\begin{center}
\includegraphics[width=1.0\textwidth]{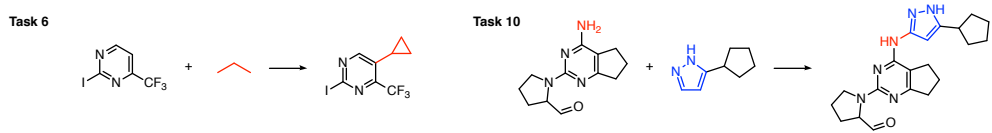}
\caption{In this AB example (task 1), the chemists preferred the literature route proceeding via a Grignard reaction, in contrast to the MCTS route, which was proposed to proceed via Seyferth-Gilbert homologation with the Ohira-Bestmann reagent. While the MCTS route is chemically reasonable, it uses less conventional chemistry in this case. 
The subsequent key steps to build the annulated cycle are the same for both MCTS and the literature.
b) Without applying the expansion policy and the in-scope filter to select the best reactions, heuristic BFS produces the typical errors traditionally criticised in retrosynthetic systems: The system tries to apply rules that are overgeneral and will not work in this molecular context.
}
\label{fig:beispiele}
\end{center}
\end{figure}

Here, one might expect that the participants can clearly identify the routes suggested by the machine as inferior. Surprisingly, this is not the case. We found that the experts did not significantly prefer the literature route (43.0\%) over our programs' route (57.0\%). Figure \ref{fig:abtest} a) shows the preferences ratios on the individual routes. Here, the preference is generally balanced, with a slight trend towards MCTS. In some cases, the participants have clear preferences (see Fig. \ref{fig:beispiele}a and Extended Methods for a examples where the chemists did not prefer MCTS).

In the second test, the participants had to report their preferences for either routes found by 3N-MCTS or routes generated by a baseline system, which uses heuristic BFS, the same transformation rules as 3N-MCTS. However, it lacks the policy network to preselect promising transformations and the in-scope filter to exclude unlikely steps. Here, the participants significantly preferred the routes generated by the MCTS algorithm (68.2\%) over the baseline system (31.8\%). We attribute the preference towards the 3N-MCTS generated routes to lower frequencies of unreasonable steps (see Figure \ref{fig:beispiele} b)).

\section*{Discussion}
In this work, we have shown that Monte Carlo Tree Search combined with deep neural network policies and an in-scope filter can be used to effectively perform chemical synthesis planning. One advantage of our approach is that it can provide full retrosynthetic pathways within seconds. 
In contrast to earlier approaches, our purely data-driven ansatz can be initially set up within a few days without the need for tedious and biased expert encoding or curation, and is equally applicable to both in-house and discipline-scale datasets. We have demonstrated that our 3N-MCTS approach has the best performance characteristics in terms of speed and number of solved problems compared with established search methods. In the past, retrosynthetic systems have been criticized for producing more noise than signal. Here, we have shown by double blind AB testing that organic chemists did not generally prefer literature routes over routes found by 3N-MCTS.
We observed that heuristic best-first search without neural network guiding did lead to many unreasonable steps being proposed in the routes, while the 3N-MCTS approach proposed more reasonable routes. This is supported by the double-blind AB experiments, where the participating organic chemists showed clear preference towards 3N-MCTS over the traditional approach. 

A number of challenges still remain, for example, investigating natural product synthesis is beyond the capabilities of our method. The sparsity of the training data in this area remains a challenge for machine learning approaches. However, natural product synthesis is also challenging for the best human chemists, as natural products can behave unpredictably, and it often requires intense methodology development.\cite{sierra2000dead} This might be solvable by stronger, but slower reasoning algorithms\cite{segler2017modelling,rocktaschel2017end} 
 Another important challenge to be solved is stereochemistry. Convincing global approaches for the quantitative prediction of enantiomeric or diastereomeric ratios without recourse to time-consuming quantum-mechanical calculations remain to be reported.\cite{peng2016computing} 
 
For the last 60 years, experts have been trying to dictate chemistry's rules to computers via hand-coded heuristics. 
After the success in game domains, for example Atari and Go, we anticipate that equipping machines with strong, general planning algorithms and the means to autonomously learn from the rich history of chemistry will finally allow them to become accepted as valuable assistants in a real-world domain which is central to solve humanity's most pressing problems in agriculture, healthcare and material science.

\section*{Acknowledgements}
M.S. and M.P.W. thank the Deutsche Forschungsgemeinschaft (SFB858) for funding. M.S. and M.P.W. would also like to thank D. Evans (RELX IntellectualProperties) and J. Swienty-Busch (Elsevier Information Systems) for the reaction dataset.  We thank all AB-test participants in Shanghai and M\"unster, and J. Guo for assistance in AB testing. M.S. thanks A. Studer, U. Hennecke, M. Wiesenfeldt, S. McAnanama-Brereton, R. Vidyadharan, and T. Kogej for valuable discussions.

\bibliography{mcts}
\bibliographystyle{naturemag}

\section*{Supporting Information}
\subsection*{Chemistry}
Molecules are treated as molecular graphs, which are vertex-labeled and edge-labeled graphs $m=(A,B)$, with atoms $a \in A$ as vertices and bonds $b \in B$ as edges. Retrosynthetic transformation rules are productions on graphs.\cite{andersen2014generic} The Chemistry Development Kit (CDK)\cite{steinbeck2006recent} and rdkit (http://www.rdkit.org) chemoinformatics libraries were used to implement the program.

\subsection*{Retrosynthesis as a Markov Decision Process}
Markov Decision Processes (MDPs) model sequential decision processes of an agent in an environment.\cite{bartosutton} An MDP is a tuple $(\mathcal{S},\mathcal{A},\mathcal{T},\mathcal{R})$, with states (positions) $s \in \mathcal{S}$, actions (transformations) $a \in \mathcal{A}$, a transition model $\mathcal{T}(s,a,s^\prime)$ determining the probability $Pr(s^\prime|a,s)$ of reaching state $s^\prime$ when taking action $a$ in state $s$ and a reward function $\mathcal{R}(s,a,s^\prime)$, which returns the reward when transitioning to $s^\prime$ via action $a$ in state $s$. A policy $\pi(a|s)$ is a probability distribution over all actions given state $s$.

Unlike in games, such as chess or go, where it is trivial to write down the ground truth rules of the game, querying the ``chemical environment'' to find out if an action actually leads to the desired successor state is expensive: Either, a wet-lab experiment has to be conducted, or a quantum-chemical calculation on a high level of theory has to be run to elucidate the reaction mechanism, which takes usually longer than running the lab experiment. Learning from millions of episodes of self-play can therefore not be employed.

To avoid these expensive interactions, we therefore need to construct a model of the environment to perform planning. As elaborated in the introduction, this model will be inaccurate.\cite{silverthesis,bartosutton} Even the best human chemists' predictions can and do fail, which entails that also humans perform synthesis planning with inaccurate ``mental environment models''. Here, we use automatic rule extraction to determine the action set (the transformations) and the in-scope filter network to learn a transition model (which is applied in a binary way). The expansion policy network serves as a prior policy.

In this work, a state (position) $s \in \mathcal{S}$ is a set of molecules $s=\{m_i,m_j,...\}$. The initial state $s_0 = \{m_0\}$ contains only the target molecule $m_0$. Actions are then transformations (rules/productions on graphs\cite{andersen2014generic}) applied to one of the molecules $m$ in a state $s$. When applying a legal action $a_l$ to a state $s_i = \{m_a,m_b,m_c\}$, it will produce a new state, e.g. $s_j = \{m_d,m_e,m_b,m_c\}$.

Given a set of building block molecules $\mathfrak{R}$ (specified before the start of the search), state $s_k$ is solved if all molecules $m_i$ in $s_k$ are building blocks $r_j \in \mathfrak{R}$. A state is terminal either if it is solved or if no legal actions are available. 
The fact that the building blocks $\mathfrak{R}$ have to be flexibly provided makes it challenging to define or learn value functions, because a changed set of building blocks changes the terminality of states and the reward function, which entails a change of the value function.\cite{bartosutton} A further challenge is that the initial state (the target molecule) changes. In most games the reward function is always the same (the rules never change, and a terminal state is always terminal). Monte Carlo Tree Search allows the calculation of value functions focused on a particular initial state on the fly,\cite{silverthesis} which makes it well suited for discovering retrosynthetic routes.

The size of the state space can only be roughly estimated. The number of drug-like molecules, which contain a restricted set of elements and functional groups, might already exceed 10\textsuperscript{60} molecules.\cite{reymond2012enumeration} However, this number excludes synthetic intermediates and organometallic and organo-main group chemistry, which add orders of magnitude to the state space size. The action space is formed by the transforms available to the system, and the legal actions are those actions that can be applied to the molecules in a state via subgraph isomorphism.
Unlike for other game artificial intelligence problems, in retrosynthesis we can limit the depth of the tree to a relatively small number (i.e. here 25) and abort the simulation as failed if synthesis is not successful within the limit. Our trees are thus wider and less deep than for other applications.

\subsection*{Monte Carlo Tree Search (MCTS)}
MCTS is a reinforcement learning approach that combines tree search with learning from simulated episodes of experience.\cite{browne2012survey} Each edge $(s,a)$ in the search tree stores the action value $Q(s,a)$, the visit count $N(s,a)$ and a prior probability $P(s,a)$ received from the expansion policy network. 

\paragraph{Selection} In the first MCTS phase, starting at the root node, the tree policy (Eq. (1)) is used to select actions. The simulation descends the tree step by step. At each step $t$, the next action $a_t$ is selected from all available actions $\mathcal{A}(s_t)$ in $s_t$ by Eq. (1), where $N(s_{t-1},a_{t-1})$ is the visit count of the state-action pair that led to the current state, and $c$ the exploration constant. 

\begin{equation}
a_t = \underset{a \in \mathcal{A}(s_t)}{\text{arg max}}\left(\frac{Q(s_t,a)}{N(s_t,a)} + c P(s_t,a) \frac{\sqrt{N(s_{t-1},a_{t-1})}}{1+N(s_t,a)}\right)
\end{equation} %

The inclusion of the prior probability $P(s,a)$ in the second term of Eq. (1) allows the system to explore the most promising lines of analysis first. With repeated visits this term decays, allowing for the exploration of other options. 
 The tree policy is applied until a leaf node or a terminal node is found. If a leaf node is visited for the second time, it is expanded. Then, all non-building block molecules $m_i \in s_t$ are processed by means of the policy-environment pipeline. The resulting state-action pairs are added to the tree as children of the leaf node, and the most probable action according to the policy network is selected for rollout. 

\paragraph{Expansion} During expansion, the state is processed once via the expansion procedure, and the reduced top50 successor states are directly added to the tree. This trick can be applied because retrosynthesis is a ``single player game'', and we do not have to fear overlooking ``killer' moves as in two-player games.

\paragraph{Evaluation by Rollout} Before starting the rollout, the state is checked first if it is terminal. A state can be terminal if it is solved. States within the tree that are already solved are called \textit{proven}.\cite{winands2008monte} States can also be terminal if no legal actions are available in that state. Terminal states are directly evaluated with the reward function. If the state is non-terminal, a rollout is started. 
During rollout, actions are sampled recursively for each molecule in the state from the top10 actions of the rollout policy until it has been deconstructed into building blocks or a maximal recursive function call depth of $d_r$ is exceeded. 

The reward function $r(s)$ returns $z>1$ if the state is proven to encourage exploitation, a reward $\in [0,1]$ depending on the ratio of molecules solved during rollout, and -1 if the state is terminal and unproven, or unsolved during rollout.

\paragraph{Update} In the update phase, the action-values $Q(s,a)$ and visit counts $N(s,a)$  of the edges traversed in the branch from $s_t$ to the root node are updated. 
The edges gather the mean action value as in Eq. \eqref{eq:indi}, where the indicator function $\mathbb{I}_i(s,a)$ is 1 if the edge was played during the $i$th simulation and $z_i$ is the reward received during rollout.

\begin{equation}
\label{eq:indi}
Q(s,a) = \frac{1}{N(s,a)} \sum_{i=1}^n \mathbb{I}_i(s,a) z_i W(b_i)
\end{equation}

Here, it is also possible to inject custom objective functions $W(b_i)\rightarrow [0,1]$, which might assign higher rewards for example to shorter, convergent, atom-economic or confident branches $b_i$. In this work, we adjust the reward based on 
\begin{gather}
\label{eq:bias}
\xi(b_i) = \text{length}(b_i) - \sum_{j=1}^J k P(s_j,a_j) \\
W^{L_\text{max}}(b_i) = \max \left( 0, \frac{L_\text{max} - \xi(b_i)}{L_\text{max}} \right)
\end{gather}
where $J=\text{length}(\cdot)$ denotes the length of a branch, $P(s_j,a_j)$ is the probability of the $j$th action in the branch obtained from the expansion policy, k=0.99 is a damping factor, and the maximal branch length $L_\text{max} = 25$. Here, inclusion of the prior policy $P(s_j,a_j)$ allows to bias also the reward towards more confident branches.

After either the time or the iteration step budget has been exhausted, the synthesis plan is selected starting at the root node by greedily choosing the action with the highest action value until a terminal solved state is reached, or a maximum depth has been exceeded, in which case the problem is unsolved.

\subsection*{Automatic transformation rule extraction}
Formalizing chemical by knowledge by hand has been attempted. Even though it sounds temptingly simple to "write down the rules of chemistry", it takes years to formalize only humble knowledge bases. Similar to rule-based common sense reasoning, this approach is considered to have exhausted its potential.\cite{ihlenfeldt1996computer,ugi1994models,kayala2011learning,kayala2012reactionpredictor} Given the exponential growth of chemical knowledge (it doubles roughly every 15 years), manual encoding is a hopeless endeavor. 

Following our previously reported procedure,\cite{neural-symbolic} building on previous work,\cite{christ2012mining,bogevig2015route,law2009route} transformation rules were therefore extracted automatically. The rules are stored using the RDkit reaction SMARTS format. 
A very general rule definition was employed for the expansion rule set, where only the atoms of the reaction center including implicit hydrogen and neighboring atom count were extracted. The rules in the rollout set contain the reaction center atoms with implicit hydrogen and neighboring atom count and additionally the directly neighboring atoms of the atoms in the reaction center with implicit hydrogen count. Rules were extracted only from reactions with one, two or three reactants and a single product. As this work is a proof-of-concept study with the intent to radically avoid expert encoding and curation, we chose not to exclude reactions based on low yield or extreme reaction conditions, as these are quite subjective criteria. For example, a yield of just a few percent can be sufficient if the aim is just to obtain a few milligrams of a compound for biological testing, while 90\% yield is clearly unsatisfying if quantitative alternatives are available. In the future, more sophisticated approaches based on reaction classification could be employed to extract rules.\cite{gelernter1990building,rose1994horace,law2009route}

The general advantages and limitations of automatic rule extraction have been discussed in detail elsewhere.\cite{ihlenfeldt1996computer,bogevig2015route,cook2012computer,law2009route,szymkuc2016computer,neural-symbolic} Its main disadvantages (defining the scope of reactions and competing reactivity; incorporating mechanistically needed/activating functional groups) can be addressed by learning supervised policies to predict the rules.\cite{neural-symbolic}

The use of symbolic rules has the preeminent advantage that it is deeply rooted in chemists' language. This  makes it easy for the model to communicate its results to the human user. Furthermore, as the transformations were extracted from the literature, this allows to directly link back to literature precedent, which is of crucial importance for users.

\subsection*{Policy networks}
The neural policy networks were trained by minimizing the negative log likelihood of selecting the transform $a$ that was used in literature to make molecule $m$. This is essentially supervised multi-class classification. Training was carried out using stochastic gradient descent within 1--2 days on a single K80 GPU. Keras was employed as the neural network framework, using Theano as the backend.\cite{chollet2015keras,team2016theano}
 
\paragraph{Expansion Policy network}
Molecules are represented by counted ECFP4 fingerprints,\cite{rogers2010extended} which are first modulo-folded to 1,000,000 dimensions, and then $\log_e(x+1)$-preprocessed. After that, a variance threshold is applied to remove rare features, leaving 32681 dimensions. As machine learning models, Highway networks with exponential linear unit (ELU) nonlinearities are used.\cite{srivastava2015training,clevert2015fast} The last layer of the neural networks is a softmax, which outputs a probability distribution over all actions (transforms) $p(a|m)$, which forms the policy (see Figure \ref{fig:neuralnet}).

\paragraph{Rollout Policy network} For rollout, molecules are represented by counted ECFP4 fingerprints,\cite{rogers2010extended} modulo-folded to 8192 dimensions, and then $\log_e(x+1)$-preprocessed. As the rollout policy, a neural network with a single hidden dense layer was used.

\subsection*{In-scope filter}
The function of the in-scope classifier is to serve as a rapid filter for the nonsensical reactions that plague rule-based systems, for example incorrect regioisomers in electrophilic aromatic substitutions. For this purpose, we chose a binary classifier, which is fast to evaluate. The investigation of more sophisticated, but slower reaction prediction approaches is left to future work.\cite{segler2017modelling,neural-symbolic,coley2017prediction,wei2016neural} For the same reason, only the product and the reaction fingerprint serve as inputs to the classifier, albeit the exclusion of conditions makes the classifier underdetermined.\cite{segler2017modelling,coley2017prediction} The inclusion of reaction conditions as another input feature would require additional search in condition space at each step, which is infeasible given the time constraints. 

\begin{figure}[htb]
\begin{center}
\includegraphics[width=0.5\textwidth]{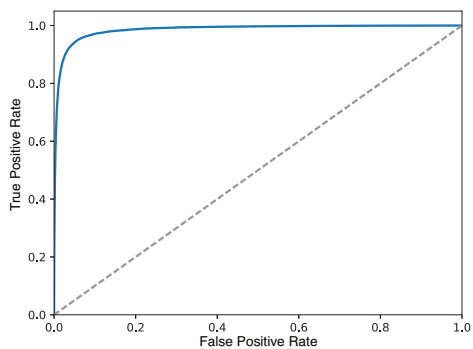}
\caption{Receiver Operation Characteristic for the in-scope filter. The area under the curve is 0.99}
\label{fig:roc}
\end{center}
\end{figure}

Our classifier is a neural network with two branches (see Figure \ref{fig:neuralnet}). The first branch embeds the reaction $r_i$, represented as a ECFP4 reaction fingerprint\cite{patel2009knowledge,zhang2005structure,landrum2014development,Polishchuk2017,marcou2015expert} $\varrho_i$ modulo-folded to 2048 dimensions, via a single dense ELU layer. The second branch embeds the product, represented as an ECFP4 fingerprint $\phi_i$, modulo-folded to 16384 dimensions and $\log_e(x+1)$-preprocessed, through a 5-layer ELU-highway network. The cosine proximity of these embeddings is then fed into a sigmoid unit to predict the probability that the reaction gives rise to the expected product. We used two strategies to obtain negative data: First, 30 million incorrect reactions were obtained by the application of reaction rules to the reactants of reported reactions.\cite{segler2017modelling,coley2017prediction} Here, we make the assumption that the reactants can only react in the reported way. Any product generated by rule application not matching the reported product is considered to be a "failed" product. With this approach for example wrong regioisomers can be generated. It is important to note that the "negative" reactions generated that way capture the cases where a naive rule-based system would fail. Furthermore, 70 million negative training datapoints were generated by perturbing tuples $(\varrho_i, \phi_i)$ to $(\varrho_i, \phi_j)$, where $i \neq j$, by negative sampling. Training data was generated only from reactions published before 2015, while test data was generated from data published in or after 2015. 
The classifier was trained by minimizing negative log likelihood using the ADAM optimizer. Figure \ref{fig:roc} (Extended Data) shows the ROC curve of the classifier. 

\subsection*{Performance Evaluation Studies}
\subsubsection*{Baselines}
In BFS, each branch is added to a priority queue, which is sorted by cost. In heuristic BFS, this cost function is the SMILES$\frac{3}{2}$ heuristic which is used as reported.\cite{szymkuc2016computer} 
In neural BFS, the cost is calculated as $f(b) = \sum_{i=0}^{s \in b}(1-P(a_i|s_i))$, where P $P(a_i|s_i))$ is the probability of that transformation calculated by the expansion policy.
Evaluation studies were performed using the CPU of a 24-core commodity cluster node using a single search thread. To provide a more meaningful comparison no GPU was used for the evaluation studies. 

It is also possible to perform the predictions in real time on a laptop (2013 MacBook Pro).

\subsection*{AB testing}
The participants in the AB tests were 45 postgraduate students specialized in organic chemistry at the Institute of Organic Chemistry at Westf\"alische Wilhelms-Universit\"at M\"unster and Shanghai Institute of Organic Chemistry.
The study was conducted in a double blind setup. During the test, neither the participants nor the conductors were aware of the origin the route.

Statistical significance was tested via the Wilcoxon signed-rank test.

\paragraph{3N-MCTS vs Literature}
In the comparison of 3N-MCTS and Literature, the expectation would be that experts prefer the literature option. In 128 ABtests, the experts preferred the literature route in 43.0\% and MCTS in 57.0\% (Wilcoxon signed-rank test on paired data, p=0.26). The null hypothesis that both data sets stem from the same source cannot be rejected.

\paragraph{3N-MCTS vs heuristic BFS}
Here, 68.2\% preferred 3N-MCTS generated solutions, while only	31.8\% preferred heuristic BFS generated solutions in 129 submitted tests. The experts significantly favor MCTS, the null hypothesis of undistinguishable sources (50\% preference for each) can clearly be rejected (Wilcoxon signed-rank test on paired data, p=0.01277).

\begin{figure}[htb]
\begin{center}
\includegraphics[width=1.0\textwidth]{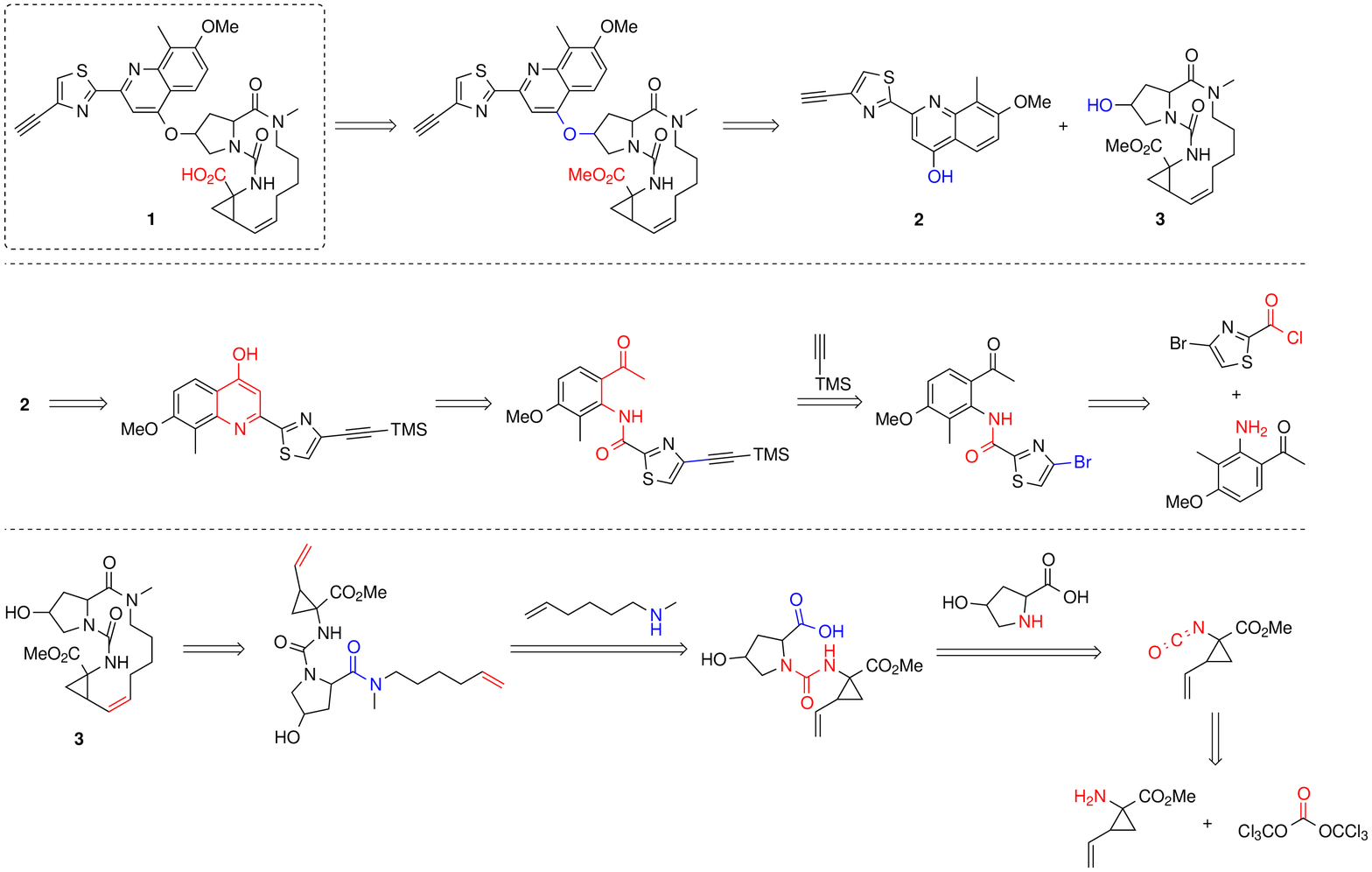}
\caption{An exemplary 10 step synthesis route for a complex intermediate in a drug synthesis.\cite{parsy2015discovery} It resembles the published route and was found by our algorithm autonomously within 30 s.}
\label{fig:example2}
\end{center}
\end{figure}

\begin{figure}[htb]
\begin{center}
\includegraphics[width=0.8\textwidth]{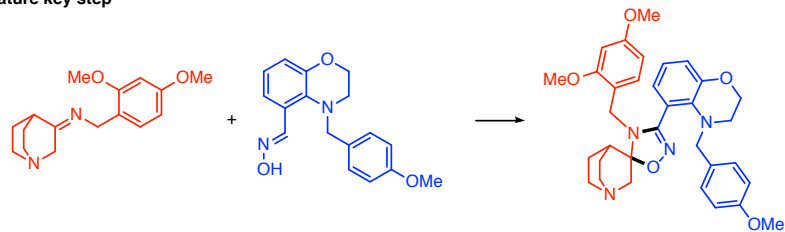}
\caption{In this task, the participants preferred the literature solution, as its key step was perceived to be more convergent.}
\label{fig:exmp3}
\end{center}
\end{figure}

\begin{figure}[htb]
\begin{center}
\includegraphics[height=0.9\textheight]{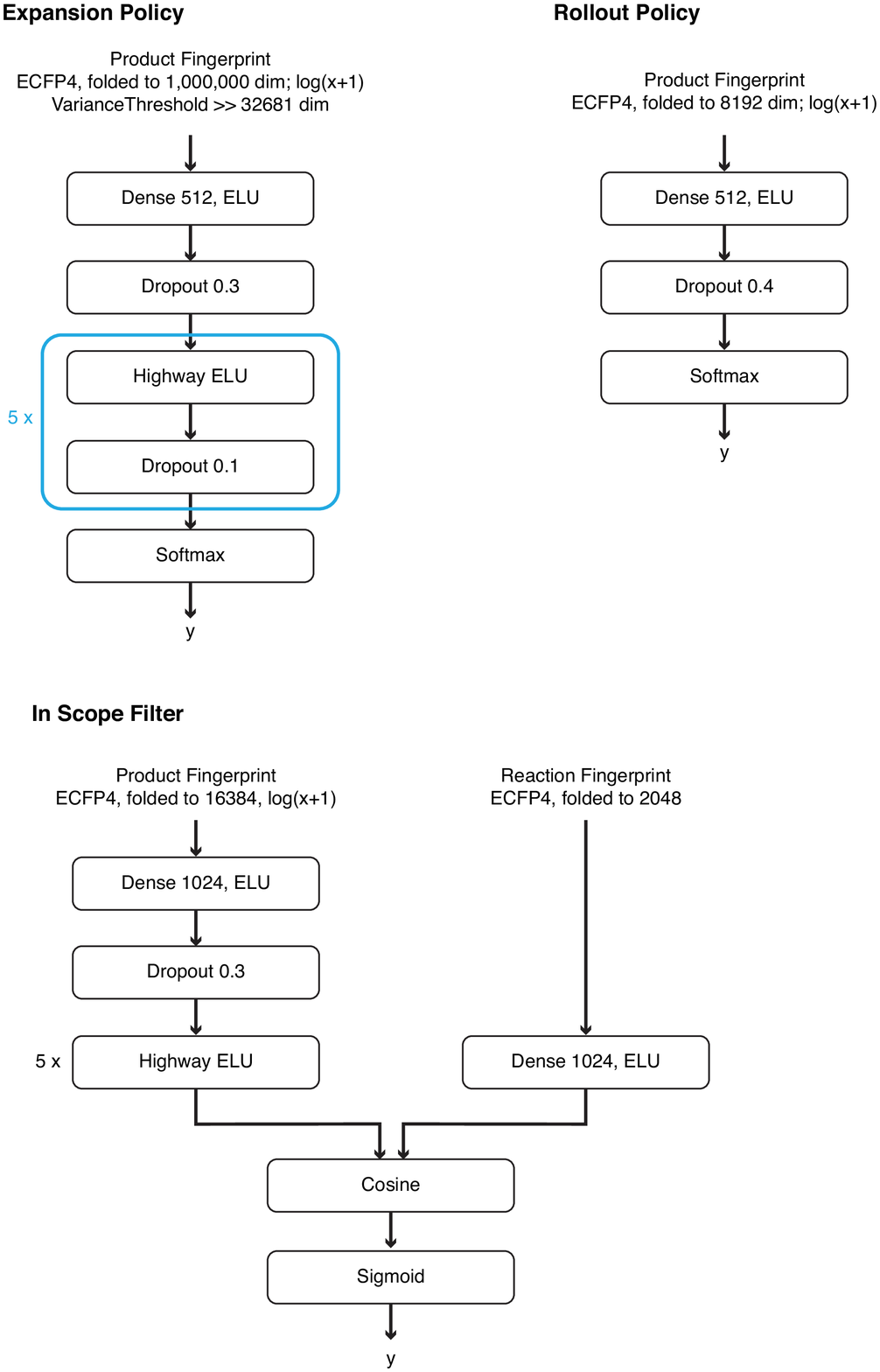}
\caption{Architectures of the used neural networks.}
\label{fig:neuralnet}
\end{center}
\end{figure}

\begin{table}[htb]
\caption{MCTS-Parameters (fine-tuned)}
\begin{center}
\begin{tabular}{lr}
\toprule
Parameter \quad & Value \\
\midrule
$d_r$ & 5\\
$c$ & 3\\
$z$ & 10\\
\bottomrule
\end{tabular}
\end{center}
\label{tab:params}
\end{table}%

\end{document}